# Using ChatGPT to Score Essays and Short-Form Constructed Responses

Mark D. Shermis

Performance Assessment Analytics, LLC

## Abstract

This study aimed to determine if ChatGPT's large language models could match the scoring accuracy of human and machine scores from the ASAP competition. The investigation focused on various prediction models, including linear regression, random forest, gradient boost, and boost. ChatGPT's performance was evaluated against human raters using quadratic weighted kappa (QWK) metrics. Results indicated that while ChatGPT's gradient boost model achieved QWKs close to human raters for some data sets, its overall performance was inconsistent and often lower than human scores. The study highlighted the need for further refinement, particularly in handling biases and ensuring scoring fairness. Despite these challenges, ChatGPT demonstrated potential for scoring efficiency, especially with domain-specific fine-tuning. The study concludes that ChatGPT can complement human scoring but requires additional development to be reliable for high-stakes assessments. Future research should improve model accuracy, address ethical considerations, and explore hybrid models combining ChatGPT with empirical methods.

# Using ChatGPT to Score Essays and Short-Form Constructed Responses

## Introduction

Twelve years ago, the Hewlett Foundation sponsored a competition (Automated Student Assessment Prize; ASAP—also referred to as the Hewlett Trials) that demonstrated the capabilities of automated essay scoring (both essays and short-form constructed responses) in an open, transparent, and replicable platform (Shermis & Hamner, 2013). The competition, which had one prize of $100,000 and another for $90,000, was divided into three parts: (1) an anonymous vendor demonstration consisting of nine commercial vendors and one university participant who scored responses from eight high-stakes essays across six different U.S. states and three grade-levels (no prize money), an open competition that consisted of 159 data-science teams from around the world who created models and scored an anonymized version of the same essays, and an open competition that consisted of 189 teams who focused on responses from ten short answer questions/prompts. All three components were administered on the Kaggle platform (http://kaggle.com), a web-based platform for data science competitions.

The competition aimed to explore whether or not high-stakes essays and short-form constructed responses processed through machine scoring algorithms predicted scores similar to those assigned by human raters. At the time, the U. S. Congress had funded assessment efforts in support of The Common Core State Standards Initiative, an effort to upgrade the K-12 curriculum in the U.S. through the use of "deeper learning" (referred to as "Race-to-the-Top"). The concern was that many of the assessments would require a substantial amount of student writing to meet their objectives. Grading this material would present considerable work and cost challenges. These assessments were led by two assessment consortia, Smarter Balanced and the Partnership for Assessment of Readiness for College and Careers (PARCC), who recruited the six states that provided essays and the three that provided short-form constructed responses for the trials. Since 2010, the popularity of Smarter Balanced and PARCC has diminished. Still, as of this writing, thirty-four states use assessments developed or acquired by the two consortia for their interim or high-stakes evaluations.

Several metrics were used to evaluate human and machine-scoring performance on the essays, including the calculation of means, standard deviations, exact agreement, exact+adjacent agreement, kappa, quadratic weighted kappa, and Pearson *r*. For a comprehensive review of the vendor demonstration and public competition results, see Shermis (2014). Of particular note was that human quadratic-weighted kappas ranged from 0.61 to 0.85 across the eight data sets and were closely followed by the machine ranges, which went from 0.60 to 0.84. The best vendor quadratic weighted kappa average across the data sets was $\kappa = .78$, which closely matched the human-human quadratic weighted kappa average. However, the averages of some vendors were significantly lower. The top performing participants in the public competition averaged slightly better on anonymized versions of the data sets, with the winning competitor average quadratic weighted kappa of .81. An internal study was conducted to assess the impact of anonymization on machine scoring of the essays. It showed only a slight drop in quadratic weighted kappa out to the third decimal place, which was not statistically significant ($p = .15$) (Shermis et al., 2015).

Human quadratic-weighted kappas averaged .89 across the short-form constructed response competition data sets. They ranged from 0.75 on Data Set 4 to 0.96 on Data Set 7, whereas the average competitor agreement was .72 across the data sets and ranged from .60 on Data Set 8 to .82 on Data Set 1. The average discrepancy between human rater performance and machine scoring performance on this measure ran from 0.06 on Data Sets 3 and 9 to 0.30 on Data Set 7, where human rater agreement was 0.96. There were significant differences in nine of the ten comparisons between human-human ratings and those obtained by machine-scoring algorithms.

The conclusion of the studies at the time was that machine scoring of essays was appropriate for low-stakes assessments and could be used as a second reader for high-stakes assessments, pending additional studies that examined scoring fairness or provided other evidence of scoring validity. While the actual quadratic weighted kappas of short-form constructed machine-scored responses approached that of those obtained with essays, they did not reach the levels of their human counterparts, which averaged $\kappa_\omega = .90$. The recommendation was to continue to develop better methods or algorithms before the technology could be used in high-stakes assessments.

More recently, the National Assessment of Educational Progress (NAEP) sponsored a competition to determine the feasibility of using machine prediction scoring algorithms to evaluate reading items for fourth- and eighth-grade reading (Shermis, 2024). The study consisted of two parts and was a set of open competitions. The first part (prompt-specific) asked competitors to score a test set of twenty items drawn from the 2017 NAEP administration. Twelve competitors were provided item materials, a randomly selected training and cross-validation set, and asked to make predictions for a test set where only the response text was provided. In the second competition (generic), five competitors were asked to score two reading items where no training or validation set was provided (only item materials). For the test set, competitors had only access to the text of the response but could draw upon information from other sources. In the first competition, the human rater performance was $\kappa_\omega = 0.91$; the scoring performance for the three competition winners was $\kappa_\omega = 0.89$, 0.88, and 0.87. In the generic competition, the winner had a $\kappa_\omega = 0.53$. Despite minor limitations, the conclusion was to begin a scoring trial run with operational items using prompt-specific models. Based on the technical reports of the winning competitors, part of the prompt-specific scoring performance improvements were attributed to improved text scoring methods such as deep neural networks, Latent Dirichlet Analysis, and BERT (Shermis, 2024).

This study aimed to determine whether large language models developed by ChatGPT can match the agreement rates obtained for human and machine-scored text addressed in the ASAP competition. Specifically, an investigation was made about which prediction models (linear regression, random forest, gradient boost, xgboost) would have the highest agreement rates with human raters. The main limitation is that evaluation metrics such as agreement rates represent only one aspect of evaluating the validity of scoring methods or algorithms. A more in-depth discussion of scoring validity for automated scoring can be found in (Bennett & Zhang, 2015; International Test Commission & Association of Test Publishers, 2022; Williamson et al., 2012). Note that some organizations are reluctant or even prohibit assessments from being scored by large language models (LLMs) because the examinee-provided data might become part of the corpus upon which the LLM is based.

Recent work has indicated that ChatGPT has the potential for scoring essays without an underlying empirical model (either prompt-specific or general). Latif and Zhai (Latif & Zhai, 2024) explored the fine-tuning of ChatGPT, specifically the GPT-3.5 model, for automatic scoring in science education. The focus was on creating individualized, engaging, and effective learning experiences through domain-specific fine-tuning ChatGPT, addressing generic AI models' limitations in educational contexts.

GPT-3.5 was fine-tuned with six complex assessment tasks in science education. These tasks included multi-label and multi-class assessment types, using responses from middle and high school students. The fine-tuned GPT-3.5 model's performance was compared with Google's pre-trained BERT model. The study used a variety of datasets and implemented a training scheme involving data cleaning, tokenization, model initialization, fine-tuning procedures, and evaluation metrics. The fine-tuning aimed to capture the nuances and intricacies of the educational domain to improve the model's scoring accuracy.

The results showed that the fine-tuned GPT-3.5 significantly outperformed the BERT model in automatic scoring tasks, demonstrating higher accuracy, particularly in multi-class tasks (approximately 9.1% better than BERT). The study concluded that fine-tuning GPT-3.5 for specific educational tasks can substantially enhance the precision and effectiveness of AI in educational assessments. The findings suggest that fine-tuned ChatGPT models hold great promise for improving automatic scoring in education, offering a scalable and replicable approach for various educational applications. The authors suggested that future research should address ethical considerations, potential biases, and the evolving role of teachers in AI-assisted education.

Wijekumar et al. (2024) reported on a study to develop and test an automated scoring model using ChatGPT to evaluate upper-elementary grade persuasive writing, specifically within the context of the We Write intervention. We Write, designed to improve persuasive writing outcomes for elementary students, integrates the Self-regulated Strategy Development (SRSD) framework with multimedia learning principles. The study aimed to provide efficient and effective scoring of summative assessments, enabling teachers to quickly assess pretests, customize instruction, and guide ongoing instruction. The automated scoring model aimed to mirror human rater scores to generate reliable writing assessment data without the subjectivity and labor costs associated with human scoring.

ChatGPT was used to score persuasive writing samples from 2200 elementary-grade students, comparing its performance with human raters. Essays were scored on a holistic scale from 0 to 6, and ChatGPT was trained using detailed prompt engineering rather than traditional fine-tuning. Various methods were employed to compare scores, including visualizations, correlations, mean differences, and agreement methods like kappa statistics. The results indicated that while ChatGPT scores tended to be more lenient and varied compared to human scores, there was a significant positive correlation between the two sets of scores. ChatGPT demonstrated high reliability, particularly with high-quality essays, though it required further refinement to align more closely with human scores across all proficiency levels.

The study concluded that ChatGPT has the potential to serve as a reliable and efficient tool for scoring elementary student writing, offering benefits in speed and scalability. However, to ensure better alignment with human scoring, particularly for middle and lower-quality essays, additional refinement of the AI model is necessary. The findings suggest that ChatGPT could complement human judgment, providing objective and consistent scores that support data-driven instruction. They recommended that future research include a more extensive and diverse sample, explore additional writing genres, and address ethical considerations such as fairness and bias in AI-assisted education.

# Method

## Participants

### Study 1

The publicly available ASAP essays were used in this study to assess essays and short-form constructed responses. Student essays (*N* = 22,029) were collected for eight different prompts representing six PARCC and SBAC) states (three PARCC states and three SBAC states) that were part of the two Race-to-the-Top assessment consortia (2015) during the 2010-2011 school year. The two consortia made the initial contact with the participating states, and ASAP began a series of negotiations that screened for their availability for participation, the type of essay, grade levels, the presence of multiple ratings, and the likelihood that the essays could be processed and validated in time for the launch of the vendor demonstration. Many states do not employ writing as part of their high-stakes assessment programs, and it obviously could not be included in the sample. The states that were chosen and cooperated had the endorsement of the consortia and asked that they not be identified in the report. In formulating the sample, an attempt was made to ensure a range of ethnicities and gender representation of students in the sample. However, these were not targeted to be representative samples for any particular population of students. Efforts were also undertaken to ensure that the types of prompts represented a range of typical writing tasks through prompts that might be similar to those anticipated in the new assessments. Responses represented a purposely heterogeneous mix in length of response. The sample was composed of essays from volunteer states and, therefore, cannot be assumed to be a genuinely representative sample of state practice. Senior representatives from both consortia indicated that the participating states were appropriate representatives of the other states in the respective consortia. However, due to the sample selection mechanism, the results here cannot be considered representative of any particular population since formal sampling designs were not employed for data collection. Instead, these are samples designed to reflect a diversity of scoring policies, demographic makeup, geographic region, prompt type, and other factors of interest to demonstrate a broad range of systems representing the state-of-the-art for machine scoring essays.

Three states were located in the Northeastern part of the U.S., two from the Midwest and one from the West Coast. Because the states provided no demographic information, student characteristics had to be estimated based on assumptions related to participating states, as displayed in Table 1. Student writers were drawn from three different grade levels (7, 8, 10), and the grade-level selection was generally a function of the testing policies of the participating states (e.g., a writing component as part of a $10^{th}$-grade exit exam), was ethnically diverse, and evenly distributed between males and females. No information was obtained regarding the characteristics of students who did not participate in the high-stakes testing programs. U.S. federal guidelines recognize that a tiny proportion of students may be excluded from the regular testing programs based on some stringent criteria.

------------------------------

Insert Table 1 About Here

------------------------------

Samples ranging in size from 1527 to 3006 were randomly selected from the data sets provided by the states and then randomly divided into three sets: a training set, a test set, and a validation set. The participants used the training set to create scoring models, except for one data set, which consisted of scores assigned by two human raters, a final or an adjudicated score (referred to as the *resolved score*), and the essay's text. The test set consisted of essay text only and was used as part of a blind test to compare different systems based on score predictions. The validation set was used for the public study only. It consisted solely of essays, being employed to verify their scoring code by the competition administrators (vendors were not required to show their source code) to prevent the vendors from over-fitting their models. The distribution of the samples was split into the following proportions: 60% training sample, 20% test sample, and 20% validation sample. The actual proportions vary slightly due to eliminating cases containing data errors. The distribution of the samples is displayed in Table 1. The specification of the training set size was predicated on what many states might employ for a pilot sample in pretesting items for a new version of their high-stakes writing assessment. Specific characteristics of the training sets are given in Table 2, and the test sets are shown in Table 3.

-------------------------------------

Insert Tables 2 & 3 About Here

-------------------------------------

Study 2

Student short-form constructed responses ($N$ =25,683) were collected for ten prompts representing one PARCC state and two Smarter-Balanced states. The short-answer responses were obtained from the state's high-stakes assessments administered during the 2010 –2011 school year, where there was a second-rater score. States use a "read behind" process to evaluate the reliability of human rater performance and take a random sample of constructed responses (proportions differ by state) to accomplish this. To the extent possible, an attempt was made to make the identity of the participating states anonymous. One state was in the northeastern United States, one from the Midwest, and one from the West Coast. Because the states provided no demographic information, student characteristics were estimated from several sources, as shown in Table 3.

------------------------------

Insert Table 3 About Here

------------------------------

Student writers were drawn from two different grade levels (8, 10), and the grade-level selection was generally a function of the testing policies of the participating states (e.g., short-answer component as part of a 10th-grade exit exam). The estimated population characteristics suggest that the sample was ethnically diverse and evenly distributed between male and female individuals. Samples ranging in size from 2,130 to 2,999 were randomly selected from the data sets provided by the states and then randomly divided into three sets: a training set, a test set (aka public leaderboard set), and a validation set (aka private leaderboard set). The vendors used the training set to create their scoring models, consisting of a score assigned by a human rater and the response text.

The distribution of the samples was split in the following approximate proportions: 60% training sample, 20% test, and 20% validation sample. The actual proportions vary slightly due to the elimination of cases containing either data errors or text anomalies. The distribution and characteristics of the training and test samples are displayed in Tables 5 and 6.

------------------------------------

Insert Tables 5 & 6 About Here

------------------------------------

Instruments

Study 1

Four essay prompts were drawn from traditional writing discourse modes (persuasive, expository, narrative). Four essay prompts were "source-based"—the questions asked in the prompt referred to a source document that students read as part of the assessment (e.g., sometimes referred to as a reading summary; cf, Weigle, 2013). In the training set, average essay lengths varied from $M = 94.39$ ($SD = 51.68$) to $M = 622.24$ ($SD = 197.08$). Traditional essays were significantly longer ($M = 354.18$, $SD$ 197.63) than source-based essays ($M = 119.97$, SD = 58.88; $t_{(13,334)} = 95.18$, $p < .05$). Testing times allotted for the high-stakes assessments ranged from 45 minutes to one hour. While there may be some debate about whether writing samples as short as 93 words constitute an "essay," these sample sizes reflect what many states define as essays in their high-stakes tests.

Five prompts employed a holistic scoring rubric; one was scored with a two-trait rubric, and two were scored with different multi-trait rubrics but reported as a composite score. The holistic scoring rubrics, as labeled by the states, were a holistic-analytic hybrid. Raters were asked to consider different dimensions of writing when making their assessments and to assign one overall score. The particular analytic dimensions were not scored. Generally, the trait rubrics focused on performance attributes for a specific audience and writing purpose. The type of rubric, scale ranges, scale means and standard deviations, and human rater agreements are reported in Table 1. Quadratic-weighted kappas ranged from 0.62 to 0.85, a typical range for human rater performance in statewide high-stakes testing programs at the time.

Study 2

Three short-answer prompts were drawn from the general sciences area, two from biology, and five from English/language arts. Eight of the ten prompts were "source dependent"—the questions asked in the prompt referred to a source document that students read as part of the assessment. In the test set, average word lengths varied from *M* = 22.62 (*SD* = 21.83) to *M* = 58.36 (SD = 23.71). The biology short-answer responses (Data Sets 5 & 6; *M* = 23.35, *SD* = 19.70) were significantly shorter than those from the other prompts (*M* = 47.70, *SD* = 25.35), $t_{(5098)} = 30.54$, *p* , < .0001. Testing times allotted for the high-stakes assessments varied depending on the context of the administration. Constructed handwritten responses were given during a 1-day period, whereas those typed into a computer had a restricted range of dates for completion. Demographic information for the short-form constructed responses is provided in Table 4.

All the prompts employed a holistic scoring rubric. Each rubric specified multiple criteria for the human rater to consider but asked the rater to make one overall score assignment. The ranges for scores in the rubric ran from 0 – 2 to 0 –3. The scale ranges, scale means, and standard deviations are reported in Table 5 for the training set. Table 6 shows the characteristics of the private leader board or cross-validation set. Human rater agreement information is also reported in Table 5 with associated data for exact agreement, kappa, and quadratic-weighted kappa. Quadratic-weighted kappas ranged from 0.75 to 0.97, a somewhat high range for human rater performance in statewide high-stakes testing programs (Kirst & Mazzeo, 1996; Massachusetts Department of Education, 2005). Because the training, public leaderboard, and private leaderboard sets were randomly selected from the entire data set provided by the state, the characteristics of the training and public leaderboard sets were similar to those of the private leaderboard.

## ChatGPT

ChatGPT is a language model developed by OpenAI that utilizes deep learning techniques to generate conversational human-like text responses. It is designed to understand and generate text based on a given prompt or series of messages, making it capable of engaging in interactive and dynamic conversations. ChatGPT is trained on a large amount of text data from the internet, allowing it to learn grammar, facts, and even some level of reasoning.

The working principle behind ChatGPT involves a two-step process: pre-training and fine-tuning. During pre-training, the model is exposed to a large corpus of text data and learns to predict the next word in a sentence. This process helps the model acquire a broad understanding of language patterns and semantic relationships. The fine-tuning stage involves training the model on more specific datasets and using reinforcement learning to fine-tune its responses to align with human-generated examples.

Despite its notable capabilities, ChatGPT does have some limitations. It tends to be sensitive to input phrasing and sometimes produces incorrect or nonsensical responses. It may also exhibit biases present in the training data. For example, suppose the model was created based on more formal writing (articles, newspaper reports, etc.). In that case, it may need to be sufficiently trained to handle student-written responses, especially at the lower grade levels. OpenAI has implemented safety mitigations to reduce harmful outputs, but there is still ongoing research to improve the system's robustness (Wu et al., 2023). Moreover, new versions of ChatGPT are released at a comparatively rapid pace. The latest available release (ChatGPT 4o) was used for this study.

## Procedure

### Studies 1 & 2

ChatGPT 4o was trained in the following manner:

1. Each dataset was analyzed separately.
2. ChatGPT was supplied with the prompt, the source material (if there was any), and the scoring rubric.
3. ChatGPT was given the training data, followed by the test dataset with only the response text and no score information (Public Leaderboard).
4. ChatGPT was asked to produce individual score predictions based on linear regression, random forest, gradient boost, and xgboost models.

Models

The models used in this study are commonly employed for predictions based on empirical data. These include linear regression, random forest, gradient boost, and xgboost. Others could have been investigated (i.e., support vector machines, neural networks, etc.) and may be utilized in the future.

Linear Regression

A linear regression model is a simple and commonly used statistical technique for modeling the relationship between a dependent variable and one or more independent variables. It assumes a linear relationship, represented by the equation $Y = \beta_0 + \beta_1X_1 + \beta_2X_2 + ... + \beta_nX_n + \epsilon$, where Y is the dependent variable, $X_1, X_2, ..., X_n$ are the independent variables, $\beta_0$ is the intercept, $\beta_1, \beta_2, ..., \beta_n$ are the coefficients representing the weights of the independent variables, and $\epsilon$ is the error term. The goal is to find the best-fitting line through the data points that minimizes the sum of the squared differences between observed and predicted values (Draper & Smith, 1998; Montgomery et al., 2021; Seber & Lee, 2012).

## Random Forest

A random forest model is an ensemble learning method that constructs multiple decision trees during training and outputs the mode of the classes (classification) or mean prediction (regression) of the individual trees. It leverages the principle of bagging, where each tree is trained on a random subset of the data with replacement, and random feature selection, where each split considers a random subset of features. This approach helps reduce overfitting and improve generalization by averaging out the errors of individual trees. Random forests are highly effective for various tasks and are robust against noise in the data (Breiman, 2001; Hastie et al., 2009; Liaw & Wiener, 2002).

## Gradient Boost Model

A gradient boost model is an ensemble technique that builds a series of weak learners, typically decision trees, sequentially. Each subsequent tree is trained to correct the errors of the previous trees by minimizing a loss function, often using gradient descent. The key idea is to optimize the model incrementally by adding trees that improve the overall model's performance. Gradient boosting is powerful for both regression and classification tasks and is known for its ability to handle complex data patterns and improve prediction accuracy, though it can be prone to overfitting if not properly regularized (Friedman, 2001; Hastie et al., 2009).

## XGBoost Model

XGBoost (Extreme Gradient Boosting) is an optimized implementation of the gradient boost model designed for speed and performance. It incorporates various regularization techniques (such as L1 and L2 regularization), efficient handling of sparse data, and parallel processing to enhance computational efficiency. XGBoost builds an ensemble of decision trees sequentially, where each tree corrects the errors of the previous ones by focusing on the residuals. It is highly customizable and includes early stopping, tree pruning, and advanced regularization. It is one of the most potent tools for structured data tasks in machine learning competitions and practical applications (Chen & Guestrin, 2016).

# Results

## Study 1

Table 7 shows the quadratic weighted kappa results for ChatGPT's scoring of essays for all four prediction models. For five of the nine data sets, QWKs were lower for the ChatGPT models than for human raters, sometimes appreciably. For example, the H1H2 agreement for data set 2a was .80, but the highest prediction model for ChatGPT was only .63. However, on four of the data sets, ChatGPT met or exceeded the QWKs for human raters. On data set five, the H1H2 QWK was 0.74, but the gradient boost model for ChatGPT achieved a QWK of .80. Except for data set 5, the linear regression models did not work particularly well. The random forest models performed slightly better. The model that worked the best overall was the gradient boost model, followed by xgboost. Xgboost might have worked better, but ChatGPT had difficulty converging on prediction solutions with the standard resources allocated to creating the models. The models eventually produced predictions when parameters or resources were reduced (e.g., from 100 estimators to 50 or 25; smaller training sample sizes, etc.). Figure 1 illustrates the performance of human raters and all four ChatGPT prediction models across the eight essay data sets.

-------------------------------

Insert Figure 1 About Here

-------------------------------

There appeared to be no discernable patterns regarding the type of essay, rubric, or relationship with human rater performance that would make ChatGPT performance better or worse. Performance did mimic humans insofar as lower QWKs were achieved when the range of scores was wider (Shermis & Hamner, 2013).

Study 2

Table 8 shows the quadratic weighted kappa results for ChatGPT's scoring of short-form constructed responses for the four prediction models. Across all ten data sets, QWKs were lower for the ChatGPT models than they were for human raters, again considerably so. ChatGPT predictions achieved QWKs in the '70s, but human rater agreements were in the range of high 80s or low-mid 90s for data sets one, five, six, and nine. The best performance was on data set six, where ChatGPT obtained $\kappa = .81$ with the gradient boost model, though H1H2 was $\kappa = .93$. As was true for the essay data sets, there was no particular relationship with the topic of the question. For these data sets, all questions were scored holistically, and the range of scores was restricted to either zero to two or zero to three. Probably because of the range of responses, the linear regression model had poor performance, while the other models had more homogenous predictions than the essay data sets. Figure 2 illustrates the performance of human raters and the four ChatGPT prediction models across the ten short-form constructed response data sets.

-------------------------------

Insert Figure 2 About Here

-------------------------------

## Discussion

Most assessment procedures aim to generate a valid score indicating current ability or performance level or generating feedback to improve performance. In the realm of writing assessment, the assessment enterprise can be expensive if it requires empirical models for the automated scoring of essays or short-form constructed responses since these currently incorporate relatively large samples upon which the scoring models are based. With the introduction of large language models, researchers have been exploring ways to employ artificial intelligence as a substitute for empirical models with varying degrees of success.

For example, Xiao et al.  (2024) utilized GPT-4 and GPT-3.5-turbo for scoring essays with various approaches, including zero-shot (i.e., one example), few-shot, and fine-tuning methods. The QWK scores for different configurations ranged from 0.67 to 0.80, indicating good agreement with human raters. The study highlighted the effectiveness of prompt engineering and retrieval-based approaches to enhance scoring accuracy, and these results are likely good enough for some types of formative feedback (the topic of that study).

Yoon (2023) developed an automated short answer grading (ASAG) model that provided analytic and final holistic scores. The study used one-shot prompting and a text similarity scoring model with domain adaptation using small manually annotated datasets with large language models to develop predictions. Four short-answer questions were drawn from the ASAP data files (data sets 1, 2, 5, & 6) to test out the model. Quadratic weighted kappa ranged from 0.67 to 0.71, which was lower than that obtained in the second part of the present study. The advantage of this approach was that significantly less training data was involved in creating a model.

Concerning model performance in the current studies, linear regression models had relatively poor performance. Still, Figures 1 & 2 suggest that random forest, gradient boosting, and extreme gradient boosting had similar results, with gradient boosting having the overall best performance. ChatGPT 4o required parameter reductions for extreme gradient boosting to produce prediction estimates. Even with fewer parameters, xgboost had predictions that resulted in QWKs close to simple gradient boosting. As ChatGPT versions evolve, one could imagine xgboost predictions matching or surpassing those of simple gradient boosting.

Regarding overall performance, ChatGPT 4o achieved higher human agreements with essays than it did for the short-form constructed responses. This result parallelled human performance during the ASAP trials (Shermis, 2014), with empirical AES models doing better for essays than short-form constructed responses. However, while ChatGPT did not perform as well as the *best* empirical models, it did perform better than a few of the commercial vendors during the vendor demonstration and many of the competitors in the open competition. At the time, the conclusion was, pending additional evidence of scoring validity (e.g., studies of scoring bias, tying the scoring process to theories of writing), automated essay scoring might be used as a second reader in high-stakes assessments.

This study did not examine ChatGPT performance on demographic (focal/reference group) differences since the ASAP data did not contain individually identifiable demographic information, but examining ChatGPT predictions for fairness is essential for helping to establish validity of its use, and there is some evidence that the writing of minorities and other marginalized groups might be under-represented (Glazko et al., 2024; Patil et al., 2024). Williamson, Xi, and Breyer (2012) provide a comprehensive set of guidelines for empirically-based automated scoring that would equally apply to predictions made with large language models used by ChatGPT. The conclusion from this study is that ChatGPT has the potential to aid in scoring essays (and, to a lesser extent, short-form constructed responses) but cannot be used as a stand-alone predictor. Our next step is to explore a hybrid predictor model that uses both ChatGPT and empirically derived models as predictors for essays and short-form constructed responses.

**Author Information**

Mark D. Shermis is principal for Performance Assessment Analytics, LLC. His area of expertise is in performance assessment in general and in automated essay scoring in particular. He was research director for the ASAP trials that demonstrated the viability of automated essay scoring for large-scale use. He can be reached at mshermis@gmail.com.

**References**

Bennett, R. E., & Zhang, M. (2015). Validity and automated scoring. In F. Drasgow (Ed.), *Testing and Technology* (pp. 142-173). Routledge. https://doi.org/10.4324/9781315871493-8

Breiman, L. (2001). Random forests. *Machine Learning*, *45*(1), 5-32. https://doi.org/10.1023/a:1010933404324

Chen, T., & Guestrin, C. (2016, August 13). Xgboost: A scalable tree boosting system. Proceedings of the 22nd ACM SIGKDD International Conference on Knowledge Discovery and Data Mining, San Francisco, CA.

Draper, N. R., & Smith, H. (1998). *Applied Regression Analysis* (Vol. 326). John Wiley & Sons. https://doi.org/10.1002/9781118625590

Friedman, J. H. (2001). Greedy function approximation: a gradient boosting machine. *Annals of Statistics*, *29*(5), 1189-1232. https://doi.org/10.1214/aos/1013203451

Glazko, K., Mohammed, Y., Kosa, B., Potluri, V., & Mankoff, J. (2024, June 3-6). Identifying and improving disability bias in GPT-based resume screening. Proceedings of the 2024 ACM Conference on Fairness, Accountability, and Transparency, Rio de Janeiro, BR.

Hastie, T., Tibshirani, R., Friedman, J. H., & Friedman, J. H. (2009). *The Elements of Statistical Learning: Data Mining, Inference, and Prediction* (Vol. 2). Springer. https://doi.org/10.1007/978-0-387-21606-5

Howell, W. G. (2015). Results of President Obama's Race to the Top [Report]. *Education Next*, *15*(4), 58+. https://link.gale.com/apps/doc/A430802559/AONE?u=anon~82408a9a&sid=googleScholar&xid=0b10b093

International Test Commission, & Association of Test Publishers. (2022). *Guidelines for Technology-Based Assessment*. Association of Test Publishers. http://www.testpublishers.org/white-papers

Kirst, M. W., & Mazzeo, C. (1996). *The Rise, Fall, and Rise of State Assessment in California, 1993-1996*. Policy Analysis for California Education (PACE). https://eric.ed.gov/?id=ED397133

Latif, E., & Zhai, X. (2024). Fine-tuning ChatGPT for automatic scoring. *Computers and Education: Artificial Intelligence*, *6*, 100210. https://doi.org/10.1016/j.caeai.2024.100210

Liaw, A., & Wiener, M. (2002). Classification and regression by randomForest. *R news*, *2*(3), 18-22. https://journal.r-project.org/articles/RN-2002-022/RN-2002-022.pdf

Massachusetts Department of Education. (2005). *2005 MCAS Technical Report*. https://iservices.cognia.org/files/MCAS/MCAS2005TechReport.pdf

Montgomery, D. C., Peck, E. A., & Vining, G. G. (2021). *Introduction to Linear Regression Analysis* (6th ed.). John Wiley & Sons. https://www.wiley.com/en-us/Introduction+to+Linear+Regression+Analysis%2C+6th+Edition-p-9781119578727

Patil, P., Kulkarni, K., & Sharma, P. (2024). Algorithmic issues, challenges, and theoretical concerns of ChatGPT. In *Applications, Challenges, and the Future of ChatGPT* (pp. 56-74). IGI Global. https://doi.org/10.4018/979-8-3693-6824-4.ch003

Seber, G. A., & Lee, A. J. (2012). *Linear Regression Analysis* (2nd ed.). John Wiley & Sons. https://doi.org/10.1002/9780471722199

Shermis, M. D. (2014). State-of-the-art automated essay scoring: Competition, results, and future directions from a United States demonstration. *Assessing Writing*, *20*, 53-76. https://doi.org/10.1016/j.asw.2013.04.001

Shermis, M. D. (2015). Contrasting state-of-the-art in the machine scoring of short-form constructed responses. *Educational Assessment*, *20*(1), 46 - 65. https://doi.org/10.1080/10627197.2015.997617

Shermis, M. D. (2024). Automated scoring for NAEP short-form constructed responses in reading. In M. D. Shermis & J. Wilson (Eds.), *Routledge International Handbook of Automated Essay Evaluation* (pp. 117-140). Routledge. http://dx.doi.org/10.4324/9781003397618-9

Shermis, M. D., & Hamner, B. (2013). Contrasting state-of-the-art automated scoring of essays. In M. D. Shermis & J. C. Burstein (Eds.), *Handbook of Automated Essay Evaluation: Current Applications and New Directions* (pp. 298 - 312). Routledge. https://doi.org/10.4324/9780203122761.ch19

Shermis, M. D., Lottridge, S., & Mayfield, E. (2015). The impact of anonymization for automated essay scoring. *Journal of Educational Measurement*, *52*(4), 419 - 436. https://doi.org/10.1111/jedm.12093

Weigle, S. C. (2013). English language learners and automated scoring of essays: Critical considerations. *Assessing Writing*, *18*(1), 85-99. https://doi.org/10.1016/j.asw.2012.10.006

Wijekumar, K. K., McKeown, D., Zhang, S., Lei, P.-W., Hruska, N., & Pirnay-Dummer, P. (2024). We Write automated scoring: Using ChatGPT for scoring in large-scale writing research projects. In M. D. Shermis & J. Wilson (Eds.), *The Routledge International Handbook of Automated Essay Evaluation* (pp. 178-194). Routledge. https://doi.org/10.4324/9781003397618-12

Williamson, D. M., Xi, X., & Breyer, F. J. (2012). A framework for the evaluation and use of automated essay scoring. *Educational Measurement: Issues and Practice*, *31*(1), 2 - 13. https://doi.org/10.1111/j.1745-3992.2011.00223.x

Wu, T., He, S., Liu, J., Sun, S., Liu, K., Han, Q. L., & Tang, Y. (2023). A brief overview of ChatGPT: The history, status quo and potential future development. *IEEE/CAA Journal of Automatica Sinica*, *10*(5), 1122-1136. https://doi.org/10.1109/jas.2023.123618

Xiao, C., Ma, W., Xu, S. X., Zhang, K., Wang, Y., & Fu, Q. (2024). From automation to augmentation: Large language models elevating essay scoring landscape. *arXiv preprint arXiv:2401.06431*. https://arxiv.org/abs/2401.06431

Yoon, S.-Y. (2023). Short answer grading using one-shot prompting and text similarity scoring model. *arXiv preprint arXiv:2305.18638*. https://arxiv.org/abs/2305.18638

Table 1. Sample Characteristics Estimated from Reported Demographics of the State for Study 1 (Essays) [adapted from (Shermis, 2014)].

| | Data Set # | | | | | | | | | | | | | | | |
|---|---|---|---|---|---|---|---|---|---|---|---|---|---|---|---|---|
| | 1 | | 2 | | 3 | | 4 | | 5 | | 6 | | 7 | | 8 | |
| State | #1 | | #2 | | #3 | | #3 | | #4 | | #4 | | #5 | | #6 | |
| Grade | 8 | | 10 | | 10 | | 10 | | 8 | | 10 | | 7 | | 10 | |
| Grade Level *N* | 42,992 | | 80,905 | | 68,025 | | 68,025 | | 71,588 | | 73,101 | | 115,626 | | 44,289 | |
| *n* | 2,968 | | 3,000 | | 2,858 | | 2,948 | | 3,006 | | 3,000 | | 2,722 | | 1,527 | |
| Training *n* | 1,785 | | 1,800 | | 1,726 | | 1,772 | | 1,805 | | 1,800 | | 1,730 | | 918 | |
| Test *n* | 589 | | 600 | | 568 | | 586 | | 601 | | 600 | | 495 | | 304 | |
| Validation *n* | 594 | | 600 | | 564 | | 590 | | 600 | | 600 | | 497 | | 305 | |
| Gender M% \| F% | 51.2 | 48.8 | 51.4 | 48.6 | 51.0 | 49.0 | 51.0 | 49.0 | 49.6 | 50.4 | 49.2 | 50.8 | 51.2 | 48.8 | 48.7 | 51.3 |
| Race % W%\| N% | 63.8 | 36.2 | 77.8 | 22.2 | 42.9 | 57.1 | 42.9 | 57.1 | 70.2 | 29.9 | 69.5 | 30.5 | 70.2 | 29.8 | 66.3 | 33.7 |
| Free/ Reduced Lunch % | 32.9 | | 40.0 | | 32.24 | | 32.2 | | 34.2 | | 34.2 | | 46.6 | | 41.3 | |

*Taken primarily from: National Center for Education Statistics, Common Core of Data (CCD), (2010). State Non-fiscal Survey of Public Elementary/Secondary Education, 2009–10, Version 1a. Washington, DC: U.S. Department of Education. This information was supplemented with state department of education website information or annual reports for each participating state.

M—Male

F—Female

W—White

N—Non-White

Table 2. Training Set Characteristics for Study 1 [adapted from (Shermis, 2014)]

| | Data Set # | | | | | | | | |
|---|---|---|---|---|---|---|---|---|---|
| | 1 | 2 | | 3 | 4 | 5 | 6 | 7 | 8 |
| *N* | 1,785 | 1,800 | | 1,726 | 1,772 | 1,805 | 1,800 | 1,730 | 918 |
| Grade | 8 | 10 | | 10 | 10 | 8 | 10 | 7 | 10 |
| Type of Essay | persuasive | persuasive | | source-based | source-based | source-based | source-based | expository | narrative |
| *M* # of Words | 366.40 | 381.19 | | 108.69 | 94.39 | 122.29 | 153.64 | 171.28 | 622.13 |
| *SD* # of Words | 120.40 | 156.44 | | 53.30 | 51.68 | 57.37 | 55.92 | 85.20 | 197.08 |
| Type of Rubric | holistic | trait (2) | | holistic | holistic | holistic | holistic | holistic* | holistic+ |
| Range of Rubric | 1-6 | 1-6 | 1-4 | 0-3 | 0-3 | 0-4 | 0-4 | 0-12 | 0-30 |
| Range of RS | 2-12 | 1-6 | 1-4 | 0-3 | 0-3 | 0-4 | 0-4 | 0-24 | 0-60 |
| *M* RS | 8.53 | 3.42 | 3.33 | 1.85 | 1.43 | 2.41 | 2.72 | 19.98 | 37.23 |
| *SD* RS | 1.54 | 0.77 | 0.73 | 0.82 | 0.94 | 0.97 | 0.97 | 6.02 | 5.71 |
| Exact Agree | 0.65 | 0.78 | 0.80 | 0.75 | 0.77 | 0.58 | 0.62 | 0.28 | 0.28 |
| κ | 0.45 | 0.65 | 0.66 | 0.61 | 0.67 | 0.42 | 0.46 | 0.17 | 0.15 |
| Quadratic Weighted κ | 0.72 | 0.81 | 0.80 | 0.77 | 0.85 | 0.75 | 0.78 | 0.73 | 0.62 |

RS-Resolved Score Adj-adjacent *composite score based on four of six traits
Agree-agreement +composite score based on six of six traits

Table 3. Test Set Characteristics for Study 1 [adapted from (Shermis, 2014)]

| | Data Set # | | | | | | | | |
|---|---|---|---|---|---|---|---|---|---|
| | 1 | 2 | | 3 | 4 | 5 | 6 | 7 | 8 |
| *N* | 589 | 600 | | 568 | 586 | 601 | 600 | 495 | 304 |
| Grade | 8 | 10 | | 10 | 10 | 8 | 10 | 7 | 10 |
| Type of Essay | persuasive | persuasive | | source-based | source-based | source-based | source-based | expository | narrative |
| *M* # of Words | 368.96 | 378.40 | | 113.24 | 98.70 | 127.17 | 152.28 | 173.48 | 639.05 |
| *SD* # of Words | 117.99 | 156.82 | | 56.00 | 53.84 | 57.59 | 52.81 | 84.52 | 190.13 |
| Type of Rubric | holistic | trait (2) | | holistic | holistic | holistic | holistic | holistic* | holistic+ |
| Range of Rubric | 1-6 | 1-6 | 1-4 | 0-3 | 0-3 | 0-4 | 0-4 | 0-12 | 0-30 |
| Range of RS | 2-12 | 1-6 | 1-4 | 0-3 | 0-3 | 0-4 | 0-4 | 0-24 | 0-60 |
| *M* RS | 8.62 | 3.41 | 3.32 | 1.90 | 1.51 | 2.51 | 2.75 | 20.13 | 36.67 |
| *SD* RS | 1.54 | 0.77 | 0.75 | 0.85 | 0.95 | 0.95 | 0.87 | 5.89 | 5.19 |
| Exact Agree | 0.64 | 0.76 | 0.73 | 0.72 | 0.78 | 0.59 | 0.63 | 0.28 | 0.29 |
| κ | 0.45 | 0.62 | 0.56 | 0.57 | 0.65 | 0.44 | 0.45 | 0.18 | 0.16 |
| Quadratic Weighted κ | 0.73 | 0.80 | 0.76 | 0.77 | 0.85 | 0.75 | 0.74 | 0.72 | 0.62 |

RS-Resolved Score Adj-adjacent *composite score based on four of six traits
Agree-agreement +composite score based on six of six traits

Table 4. Sample Characteristics Estimated from Reported Demographics of the State for Study 2 (Short-Form Constructed Responses) [adapted from (Shermis, 2015)].

| | Data Set # | | | | | |
|---|---|---|---|---|---|---|
| | 1-4 | | 5-9 | | 10 | |
| State | #1 | | #2 | | #3 | |
| Grade | 10 | | 10 | | 8 | |
| Grade Level *N* | 44,485 | | 81,245 | | 78,902 | |
| Total *n* | 10,966 | | 11,983 | | 2,734 | |
| Training *n* | 6,579 | | 7,189 | | 1,640 | |
| Test *n* | 2,195 | | 2,395 | | 546 | |
| Validation *n* | 2,192 | | 2,399 | | 548 | |
| Gender M% \| F% | 51.2 | 48.8 | 51.4 | 48.6 | 51.5 | 48.5 |
| Race % W%\| N% | 63.8 | 36.2 | 77.8 | 22.2 | 61.3 | 38.7 |
| Free/ Reduced Lunch % | 32.9 | | 40.0 | | 43.7 | |

Table 5. Training Set Characteristics for Study 2 [adapted from (Shermis, 2015)].

| | Data Set # | | | | | | | | | |
|---|---|---|---|---|---|---|---|---|---|---|
| | 1 | 2 | 3 | 4 | 5 | 6 | 7 | 8 | 9 | 10 |
| *N* | 1,672 | 1,278 | 1,891 | 1,738 | 1,795 | 1,797 | 1,799 | 1,799 | 1,798 | 1,640 |
| Grade | 10 | 10 | 10 | 10 | 10 | 10 | 10 | 10 | 10 | 8 |
| *M* # of Words | 47.12 | 59.17 | 47.78 | 40.38 | 25.06 | 23.38 | 41.13 | 53.05 | 49.73 | 41.41 |
| *SD* # of Words | 22.07 | 22.53 | 14.45 | 15.54 | 22.08 | 21.83 | 24.80 | 32.70 | 36.16 | 29.02 |
| Type of Rubric | holistic science | holistic science | holistic English-L.A. | holistic English-L.A. | holistic biology | holistic biology | holistic English-L.A. | holistic English-L.A. | holistic English-L.A. | holistic English-L.A. |
| Range of Rubric | 0-3 | 0-3 | 0-2 | 0-2 | 0-3 | 0-3 | 0-2 | 0-2 | 0-2 | 0-2 |
| Range of Score | 0-3 | 0-3 | 0-2 | 0-2 | 0-3 | 0-3 | 0-2 | 0-2 | 0-2 | 0-2 |
| *M* Score | 1.49 | 1.73 | 0.99 | 0.69 | 0.29 | 0.25 | 0.71 | 1.13 | 1.10 | 1.18 |
| *SD* Score | 1.05 | 0.98 | 0.69 | 0.61 | 0.60 | 0.66 | 0.82 | 0.85 | 0.76 | 0.71 |
| Exact Agree | 0.89 | 0.85 | 0.76 | 0.78 | 0.97 | 0.97 | 0.96 | 0.84 | 0.81 | 0.88 |
| κ | 0.86 | 0.80 | 0.60 | 0.61 | 0.91 | 0.89 | 0.93 | 0.75 | 0.71 | 0.82 |
| Quadratic Weighted κ | 0.94 | 0.91 | 0.73 | 0.70 | 0.95 | 0.96 | 0.97 | 0.86 | 0.83 | 0.88 |

Table 6. Test Set Characteristics for Study 2 [adapted from (Shermis, 2015)].

| | Data Set # | | | | | | | | | |
|---|---|---|---|---|---|---|---|---|---|---|
| | 1 | 2 | 3 | 4 | 5 | 6 | 7 | 8 | 9 | 10 |
| *N* | 558 | 426 | 318 | 250 | 599 | 599 | 601 | 601 | 600 | 548 |
| Grade | 10 | 10 | 10 | 10 | 10 | 10 | 10 | 10 | 10 | 8 |
| *M* # of Words | 48.47 | 58.36 | 48.51 | 38.62 | 22.62 | 24.08 | 40.96 | 54.18 | 51.33 | 41.13 |
| *SD* # of Words | 21.92 | 23.71 | 15.10 | 16.26 | 18.14 | 21.26 | 24.82 | 33.53 | 40.09 | 27.33 |
| Type of Rubric | holistic science | holistic science | holistic English-L.A. | holistic English-L.A. | holistic biology | holistic biology | holistic English-L.A. | holistic English-L.A. | holistic English-L.A. | holistic English-L.A. |
| Range of Rubric | 0-3 | 0-3 | 0-2 | 0-2 | 0-3 | 0-3 | 0-2 | 0-2 | 0-2 | 0-2 |
| Range of Score | 0-3 | 0-3 | 0-2 | 0-2 | 0-3 | 0-3 | 0-2 | 0-2 | 0-2 | 0-2 |
| *M* Score | 1.53 | 1.71 | 0.98 | 0.68 | 0.31 | 0.27 | 0.76 | 1.16 | 1.11 | 1.22 |
| *SD* Score | 1.01 | 1.03 | 0.67 | 0.65 | 0.65 | 0.67 | 0.83 | 0.85 | 0.78 | 0.68 |
| Exact Agree | 0.90 | 0.85 | 0.80 | 0.82 | 0.96 | 0.95 | 0.96 | 0.85 | 0.81 | 0.88 |
| $\kappa$ | 0.86 | 0.80 | 0.60 | 0.61 | 0.91 | 0.89 | 0.93 | 0.75 | 0.71 | 0.82 |
| Quadratic Weighted $\kappa$ | 0.95 | 0.93 | 0.77 | 0.75 | 0.95 | 0.93 | 0.96 | 0.86 | 0.84 | 0.87 |

Table 7. Results from ChatGPT 4-o Prediction on ASAP Essay Data Sets

| Dataset # | Training n | Test n | Training Mean | Test Mean | Training SD | Test SD | H1H1 κ | H1GPT LR κ | H1GPT RF κ | H1GPT GB κ | H1GPT XGB κ |
|---|---|---|---|---|---|---|---|---|---|---|---|
| 1 | 1785 | 589 | 8.53 | 8.52 | 1.54 | 1.54 | 0.73 | 0.63 | 0.64 | 0.73 | 0.72 |
| 2a | 1800 | 600 | 3.42 | 3.41 | 0.77 | 0.77 | 0.80 | 0.58 | 0.56 | 0.63 | 0.54 |
| 2b | 1800 | 600 | 3.33 | 3.32 | 0.73 | 0.75 | 0.76 | 0.45 | 0.48 | 0.56 | 0.51 |
| 3 | 1726 | 568 | 1.85 | 1.90 | 0.82 | 0.85 | 0.77 | 0.49 | 0.53 | 0.54 | 0.58 |
| 4 | 1772 | 586 | 1.43 | 1.51 | 0.94 | 0.95 | 0.85 | 0.60 | 0.66 | 0.72 | 0.67 |
| 5 | 1805 | 601 | 2.41 | 2.51 | 0.97 | 0.95 | 0.74 | 0.79 | 0.71 | 0.80 | 0.78 |
| 6 | 1800 | 600 | 2.72 | 2.75 | 0.97 | 0.87 | 0.74 | 0.67 | 0.73 | 0.78 | 0.77 |
| 7 | 1730 | 495 | 19.98 | 20.13 | 6.02 | 5.89 | 0.73 | 0.49 | 0.44 | 0.51 | 0.51 |
| 8 | 918 | 304 | 37.23 | 36.67 | 5.71 | 5.19 | 0.62 | 0.57 | 0.60 | 0.65 | 0.65 |

LR-Linear Regression RF-Random Forest GB-Gradient Boosting XGB-XGB Boost

For most XGB models, Chat_GPT 4o had to reduce the number of estimators, sample size of the training set, depth of trees, or a combination of factors in order to successfully build the model.

Table 8. Results from ChatGPT 4-o Prediction on ASAP Short-Form Constructed Response Data Sets

| Dataset # | Training n | Test n | Training Mean | Test Mean | Training SD | Test SD | H1H1 κ | H1GPT LR κ | H1GPT RF κ | H1GPT GB κ | H1GPT XGB κ |
|---|---|---|---|---|---|---|---|---|---|---|---|
| 1 | 1,672 | 558 | 1.49 | 1.53 | 1.05 | 1.01 | 0.95 | 0.20 | 0.71 | 0.74 | 0.71 |
| 2 | 1,278 | 426 | 1.73 | 1.71 | 0.98 | 1.03 | 0.93 | 0.30 | 0.46 | 0.52 | 0.40 |
| 3 | 1,891 | 406 | 0.98 | 0.92 | 0.66 | 0.67 | 0.76 | 0.30 | 0.53 | 0.57 | 0.59 |
| 4 | 1,738 | 579 | 0.69 | 0.68 | 0.61 | 0.65 | 0.75 | 0.49 | 0.58 | 0.58 | 0.56 |
| 5 | 1,795 | 598 | 0.29 | 0.31 | 0.60 | 0.65 | 0.95 | 0.10 | 0.70 | 0.76 | 0.73 |
| 6 | 1,797 | 599 | 0.25 | 0.27 | 0.66 | 0.67 | 0.93 | 0.47 | 0.80 | 0.81 | 0.77 |
| 7 | 1,799 | 599 | 0.71 | 0.76 | 0.82 | 0.83 | 0.96 | 0.17 | 0.58 | 0.57 | 0.54 |
| 8 | 1,799 | 599 | 1.13 | 1.16 | 0.85 | 0.85 | 0.86 | 0.23 | 0.48 | 0.50 | 0.47 |
| 9 | 1,798 | 600 | 1.10 | 1.11 | 0.76 | 0.78 | 0.84 | 0.39 | 0.74 | 0.72 | 0.66 |
| 10 | 1,640 | 546 | 1.18 | 1.22 | 0.71 | 0.68 | 0.87 | 0.20 | 0.70 | 0.62 | 0.60 |

LR-Linear Regression  RF-Random Forest  GB-Gradient Boosting XGB-XGB Boost

For most XGB models, Chat_GPT 4o had to reduce the number of estimators, sample size of the training set, depth of trees, or a combination of factors in order to successfully build the model.

Figure 1. Quadratic Weighted Kappas for Essay Score Predictions.

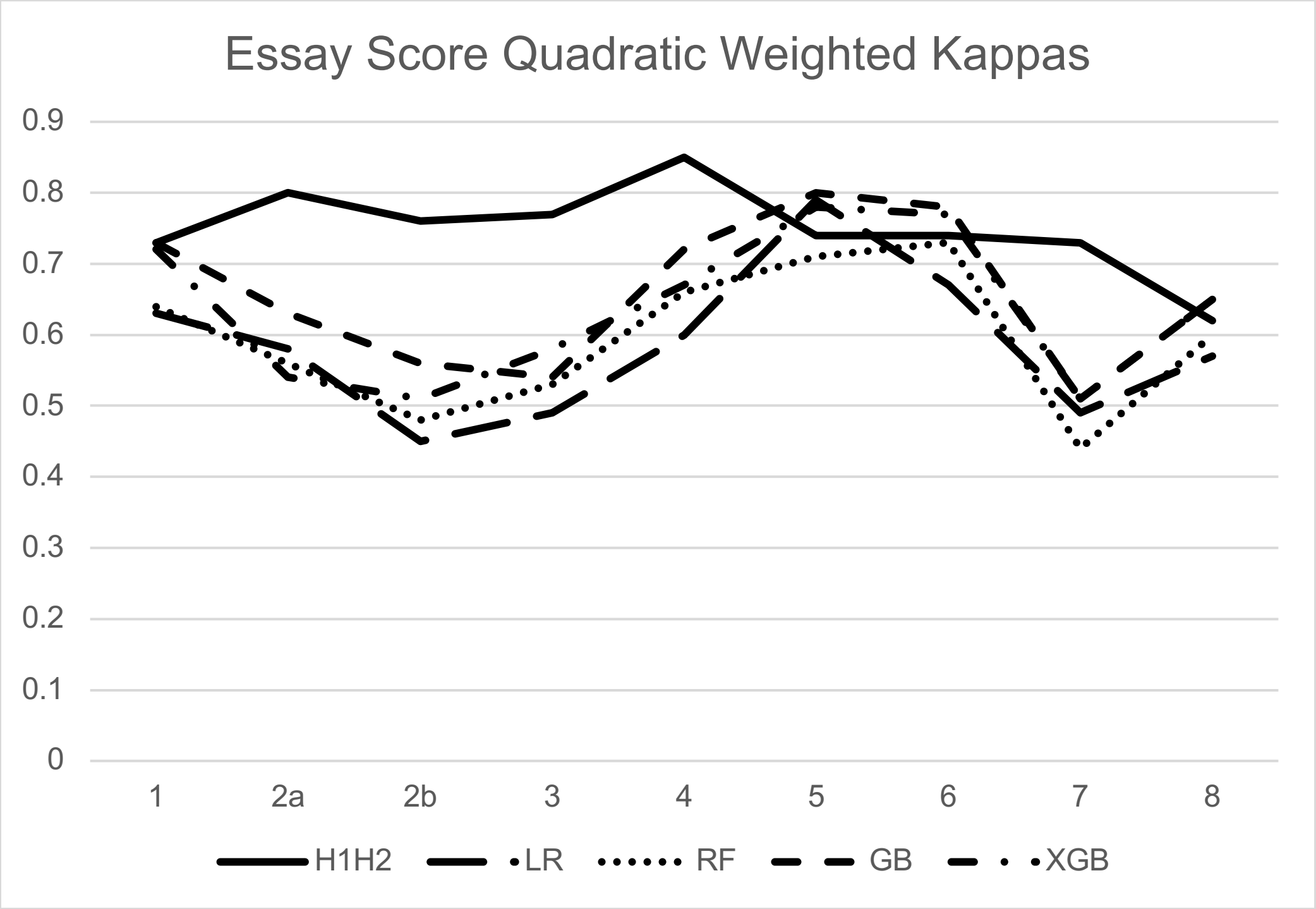

LR-Linear Regression RF-Random Forest GB-Gradient Boosting XGB-XGB Boost

Figure 2. Quadratic Weighted Kappas for Short-Form Constructed Responses Predictions.

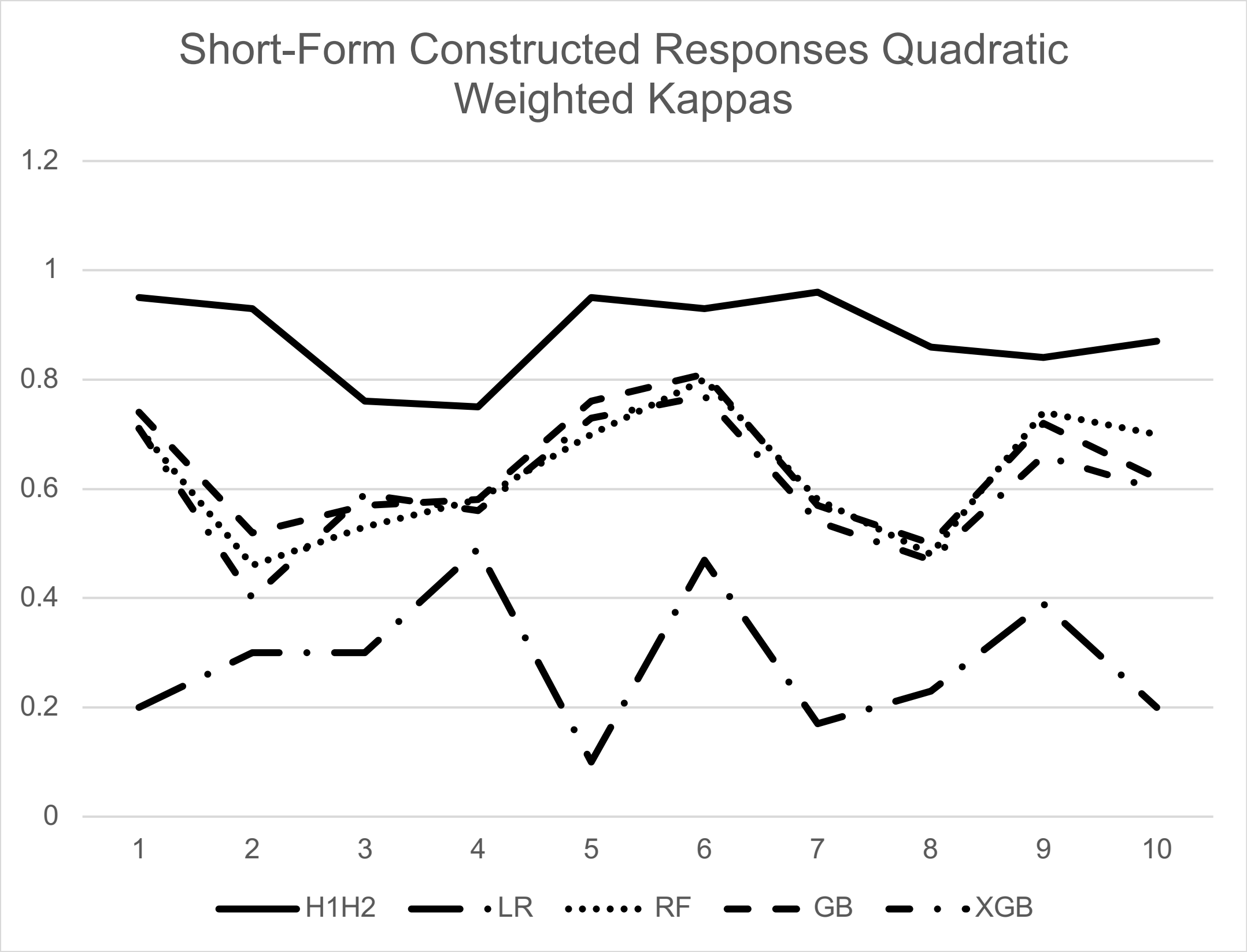

LR-Linear Regression  RF-Random Forest  GB-Gradient Boosting XGB-XGB Boost